\documentclass{article}
\usepackage[preprint]{corl_2023} 
\usepackage{graphicx}
\title{OASim: an Open and Adaptive Simulator based on Neural Rendering for Autonomous Driving}

\author{
  Guohang Yan, Jiahao Pi, Jianfei Guo, Zhaotong Luo, Min Dou, Nianchen Deng, Qiusheng Huang\\ 
  \textbf{Daocheng Fu, Licheng Wen, Pinlong Cai, Xing Gao, Xinyu Cai, Bo Zhang, Xuemeng Yang}\\
  \textbf{Yeqi Bai, Hongbin Zhou, Botian Shi\thanks{ Corresponding author.}}
  \\
  \\Shanghai AI Laboratory, Shanghai, China
}
\begin{document}
\maketitle


\begin{abstract}
    With deep learning and computer vision technology development, autonomous driving provides new solutions to improve traffic safety and efficiency. The importance of building high-quality datasets is self-evident, especially with the rise of end-to-end autonomous driving algorithms in recent years. Data plays a core role in the algorithm closed-loop system. However, collecting real-world data is expensive, time-consuming, and unsafe. With the development of implicit rendering technology and in-depth research on using generative models to produce data at scale, we propose OASim, an open and adaptive simulator and autonomous driving data generator based on implicit neural rendering. It has the following characteristics: (1) High-quality scene reconstruction through neural implicit surface reconstruction technology. (2) Trajectory editing of the ego vehicle and participating vehicles. (3) Rich vehicle model library that can be freely selected and inserted into the scene. (4) Rich sensors model library where you can select specified sensors to generate data. (5) A highly customizable data generation system can generate data according to user needs. We demonstrate the high quality and fidelity of the generated data through perception performance evaluation on the Carla simulator and real-world data acquisition. Code is available at \url{https://github.com/PJLab-ADG/OASim}.
\end{abstract}



\section{Introduction}

    Autonomous driving~\citep{hu2023planning, caesar2020nuscenes, chen2023end} is a rapidly advancing technology that holds the promise of a future where commuting is safer, faster, and more convenient. However, the large-scale deployment of autonomous vehicles has been hindered by several challenges, chief among them being the need for vast amounts of data~\citep{liu2020overview}. To cope with the complex real-world environment and ensure the safety of drivers and passengers on road, large amount of data is required for rigorous and thorough testing~\citep{kaur2021survey}. Besides, the deep learning methods that have the major contribution to the remarkable progress of autonomous driving are hungry for diverse data-sets and data-tags. 
    
    
    Developing and testing algorithms in real world, although an integral part of the vehicle development process, is expensive, time-consuming, and has potential safety risks and legal restrictions. Moreover, the data collected on actual road will inevitably be duplicated and redundant. As the algorithms become more capable, the effectiveness of road collected data will drop significantly and finding corner cases will be extremely difficult. 

    Simulation is a low cost, efficient and customized way to test new concepts, strategies, and algorithms. It has been used since the very beginning of autonomous driving research~\citep{nelson1988local, pomerleau1988alvinn}. A lot of works are devoted to the design of physical engines~\citep{hugues2006simbad, hu2019chainqueen}, simulating sensor observations~\citep{manivasagam2020lidarsim, fang2020augmented} and constructing the 3D environment. Early works focused on accurate modelling of the kinematic and dynamics behaviours of a vehicle~\citep{koenig2004design, benekohal1988carsim}. However, the environment they create is simple and not close to the real world. More recently, some researchers took advantage of racing simulators~\citep{wymann2000torcs, chen2015deepdriving} and commercial games~\citep{richter2016playing, richter2017playing} to test the autonomous driving algorithms. However, the close-source commercial software and the lack of traffic rules and driving policies limits its customize development.


    
    Simulators based on game engines like Unreal Engine or Unity offer more physically and visually realistic data by performing rasterization or ray-tracing~\citep{parker2010optix}. Representative platforms include CARLA~\citep{dosovitskiy2017carla}, AirSim~\citep{dosovitskiy2017carla} and DeepDrive. However, they use artificially designed environment and assets that require expensive manual effort. There is also a sim-to-real gap caused by the unsatisfactory low photorealism and lack of diverse traffic participants.

    Instead of game-engine-based simulators, data-driven simulators which generate synthetic data directly from real data have emerged and achieved high-fidelity as well as low labor cost. AADS~\citep{li2019aads} generated traffic flows from the acquired real-world trajectory to augment simulation pictures. VISTA~\citep{amini2020learning, amini2022vista} synthesizes novel viewpoints by image-based warping or ray-casting. However, these approaches have visual artifacts, resulting in poor picture quality. The advent of Neural radiance fields(NeRFs) offers a more realistic rendering ability by using multi-layer perceptrons(MLPs) to represent a static scene. However challenges arise when it is applied to the unbounded outdoor scenes with dynamic objects. Recent approaches model the foreground instances and background environments separately with independent networks~\citep{yang2023unisim, wu2023mars}.  UniSim~\citep{yang2023unisim} uses a sparse voxel-based representation to efficiently simulate both image and LiDAR observations under a unified framework. MARS~\citep{wu2023mars} provides a modular simulator that can switch between different modern NeRF-related methods.
    
    

    
    In this paper, this technical report introduces OASim - an open and adaptive simulator for autonomous driving based on neural implicit reconstruction and rendering. The environment is represented by implicit surface reconstruction technique~\citep{guo2023streetsurf}. The static environment and moving objects are modeled separately by the network. The moving objects can form a foreground asset library that is available to be inserted to the scene. The agent appearance, driving trajectory, and sensor configuration can be customized to generate high-fidelity data for downstream applications such as perception and control. We provide a user-friendly interactive visualization interface to import and export data, edit settings, and visualize rendering results. The core features of OASim are listed below: 
    
    \begin{itemize}
        \item High-fidelity real-time rendering performance on neural implicit reconstruction.
        \item The trajectories of the ego vehicle and other actors can be edited and the traffic flow with vehicle interactions will be simulated. 
        \item Different sensor configurations can be selected for data generation.
        \item Based on the newly generated vehicle trajectory or new sensor configuration, user-defined data can be rendered.
    \end{itemize}



\section{Methodolody}
 
    \begin{figure}[htp]
        \centering
        \includegraphics[width=0.9\linewidth]{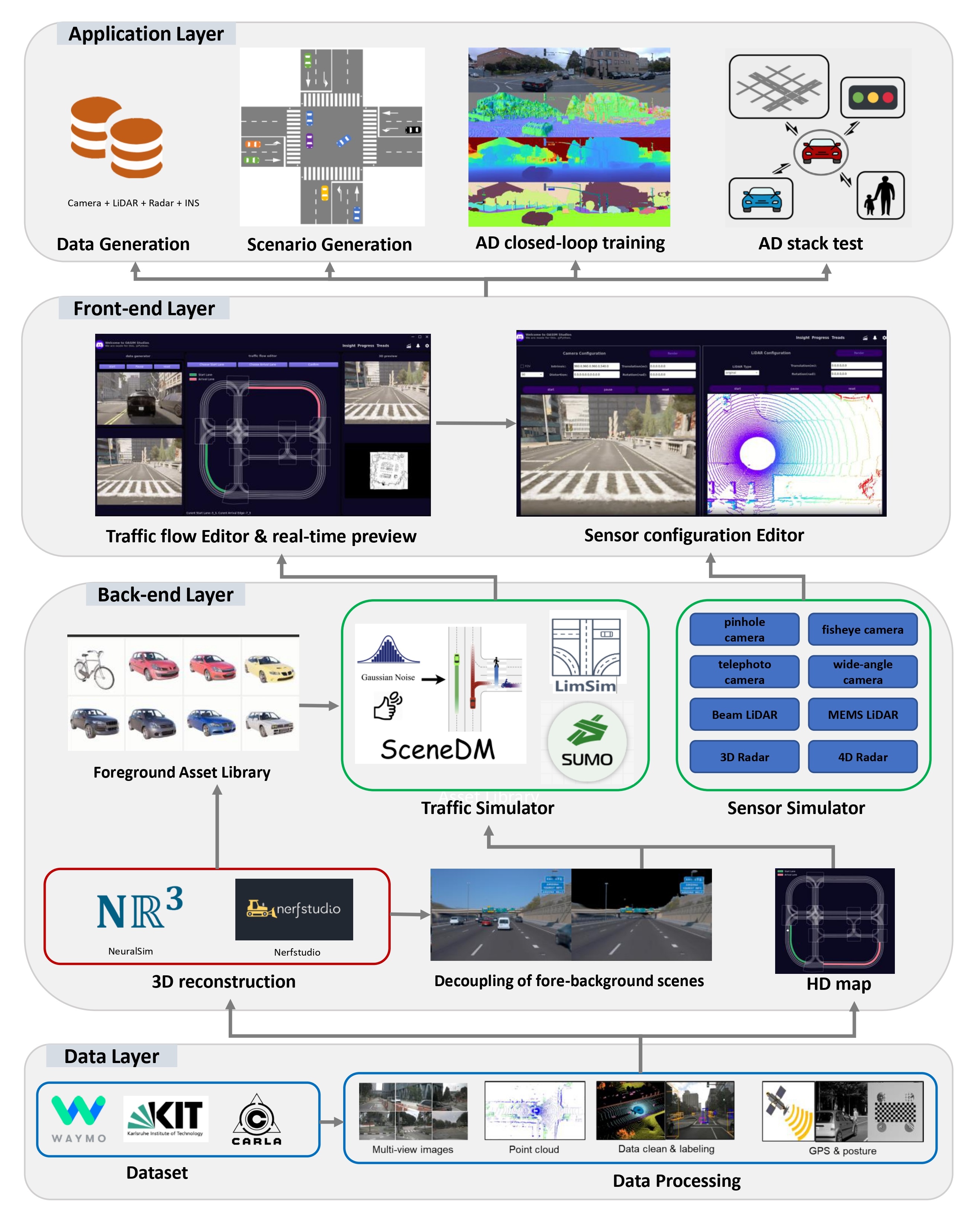}
        \caption{Workflow of OASim. Its hierarchical structure can be divided into four layers, including data layer, back-end layer, front-end layer and application layer.}
        \label{fig:framework}
    \end{figure}

    OASim focuses on generating high-fidelity and customizable autonomous driving data through neural implicit reconstruction and rendering techniques. The pipeline of OASim is shown as Fig.\ref{fig:framework}. The hierarchical structure can be divided into four layers. The first data layer converts the input data into the format we require, including data cleaning and labeling. The processed sensory data and labeled HD map are then input into the back-end layer. This layer is the core of the system and implements 3D reconstruction, traffic flow simulation and novel data synthesis. The front-end layer provides an interactive interface for users to conveniently change the vehicle route and sensor configurations. The newly synthesized data can be used for multiple downstream tasks such as perception, planning, etc.

\subsection{Data Collection and Processing}

    The real-world data can be collected by a sensor-equipped vehicle driving over a period of time. The raw data should contain RGB-images from multi-view cameras, sensor poses from IMU/GPS devices and optionally point cloud from LiDAR. Accurate calibration parameters are needed including time synchronization and spatial alignment to fuse sensor observations. 

    The data source of the system includes well-annotated public dataset like Waymo~\citep{sun2020scalability}, Nuscenes~\citep{caesar2020nuscenes}, Kitti~\citep{geiger2013vision}. The system can also be applied on custom data collected from real world or from other simulation platforms~\cite{dosovitskiy2017carla}. 
    
    The quality of the collected data greatly affects the results of reconstruction and rendering. Since the observations of self-driving vehicle are sparse and often captured from constrained viewpoints, it is recommended that the vehicle be equipped with multiple sensors at different direction and travel on different routes in a certain area.

    To accommodate the different format and unstable quality of collected data, we first convert it into a unified format and check its integrity. In order to reconstruct the static environment, we generate the dynamic object masks by segmentation or objection detection methods. While the extrinsic is not reliable enough, we use the differentiable property of the scene representation relative to the pose to jointly optimize the time offset and extrinsic parameters \citep{herau2023moisst}. Besides, we annotate the road elements such as lane-level geometry, road marking, traffic signs, and barriers manually to generate the HD map.
     
    The processed sensory information is then fed into the neural field for environment reconstruction. The labeled HD map is used in the traffic flow simulator for generating vehicle interaction and reasonable trajectory.
	
\subsection{Environment Reconstruction and Editable Rendering}
In implicit rendering, NeRF (Neural Radiance Fields) and 3D Gaussian Splatting are both advanced technologies used for reconstructing and rendering three-dimensional scenes, each with their own unique characteristics and advantages. The core idea of NeRF is to use a neural network to model the radiance and depth of every point in the scene by training a Multi-Layer Perceptron (MLP). It's trained to map spatial coordinates and viewing directions to color and density values. The advantage of this method lies in its ability to capture complex scene details through training data, thereby achieving high-quality rendering results. The rendering process of NeRF involves a volumetric representation of the scene, meaning points are sampled uniformly in three-dimensional space and their radiative properties are predicted by the network.

3D Gaussian Splatting is a point-based rendering method that uses millions of 3D Gaussian distributions to represent the geometry of shapes within a scene. Unlike NeRF, 3D Gaussian Splatting provides an explicit representation of the scene, and its rendering algorithm is differentiable, meaning it can be optimized and edited more easily. This method is characterized by combining real-time rendering capabilities with a high degree of scene control, making it a powerful tool in fields such as virtual reality and interactive media. The strength of 3D Gaussian Splatting lies in its ability to maintain high-quality reconstruction while also integrating into traditional rasterization rendering pipelines, allowing for faster optimization.

Currently, some works in autonomous driving are based on implicit reconstruction and foreground editing, such as \citep{yang2023unisim, zhou2023drivinggaussian, guo2023streetsurf}, etc. We are currently mainly using StreetSurf \cite{guo2023streetsurf} as the basis for implicit reconstruction and rendering. Meanwhile, we have also built a rich foreground asset library, where users can customize vehicles in the scene. The official open-source counterpart to StreetSurf is NeuralSim \citep{neuralsim2023}. It's important to note that, for now, NeuralSim has only released the background rendering code as open-source, with plans to release the combined foreground and background rendering code in the future. Consequently, the current open-source version of OASim only supports background rendering and will be updated as NeuralSim progresses with its open-sourcing efforts. Moving forward, we plan to incorporate more implicit reconstruction methods into the OASim system, such as DrivingGaussian \cite{zhou2023drivinggaussian} and other methods based on 3D Gaussian Splatting, and we will also enrich the variety of foreground assets, for example, by adding categories like bicycles and pedestrians.

\subsection{Interactive Visualization Interface}

\begin{figure}[htp]
    \centering
    \includegraphics[width=0.9\linewidth]{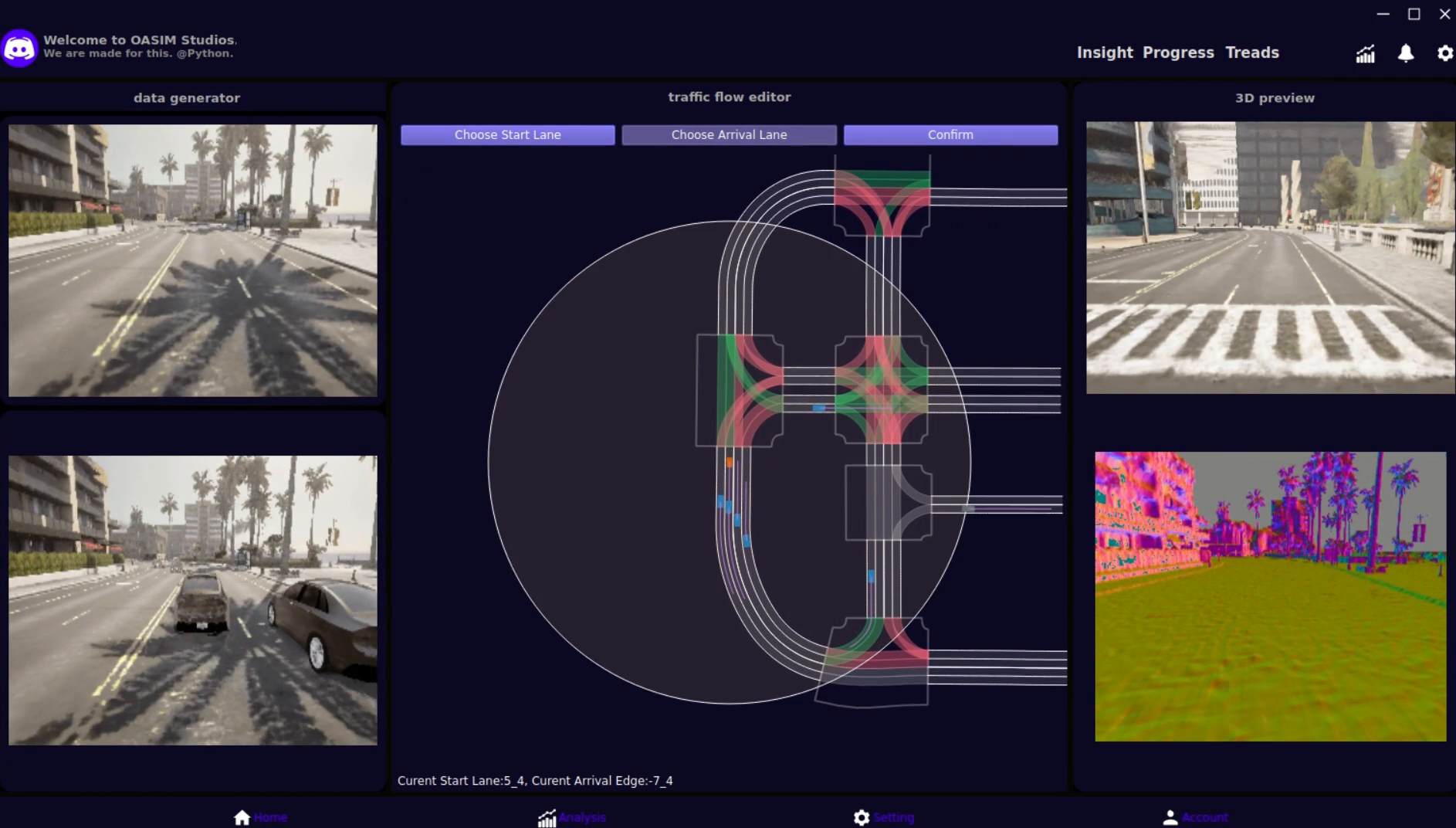}
    \caption{OASim interface. }
    \label{fig:inteface1}
\end{figure}

OASim allows users to edit the trajectory of self-vehicle and other vehicles in the scene through an interactive interface as shown in Fig.\ref{fig:inteface1}. The established road network is displayed on the middle interface to facilitate users to select vehicle routes. The trajectory can be automatically generated by selecting the starting point and destination. Alternatively, users can use keyboard controls to subtly guide the self-driving vehicle, giving detailed driving instructions such as lane changes, turns, etc. The render module can generate data according to the edited trajectory in real time. Besides, in the other widgets, users can preview the scene in different perspectives by keyboard control.

OASim allows flexible configuration of the agent's sensor suite, including cameras, LiDAR, Radar, etc. We support modifying the sensor model by changing the intrinsic and extrinsic parameters. The position and orientation of the sensor relative to the vehicle body are represented by extrinsic parameters. Some commonly used intrinsic combinations are preset in the system to facilitate user selection. Once the sensor configuration is complete, it will be used to generate and preview data.
\begin{figure}[htp]
    \centering
    \includegraphics[width=0.9\linewidth]{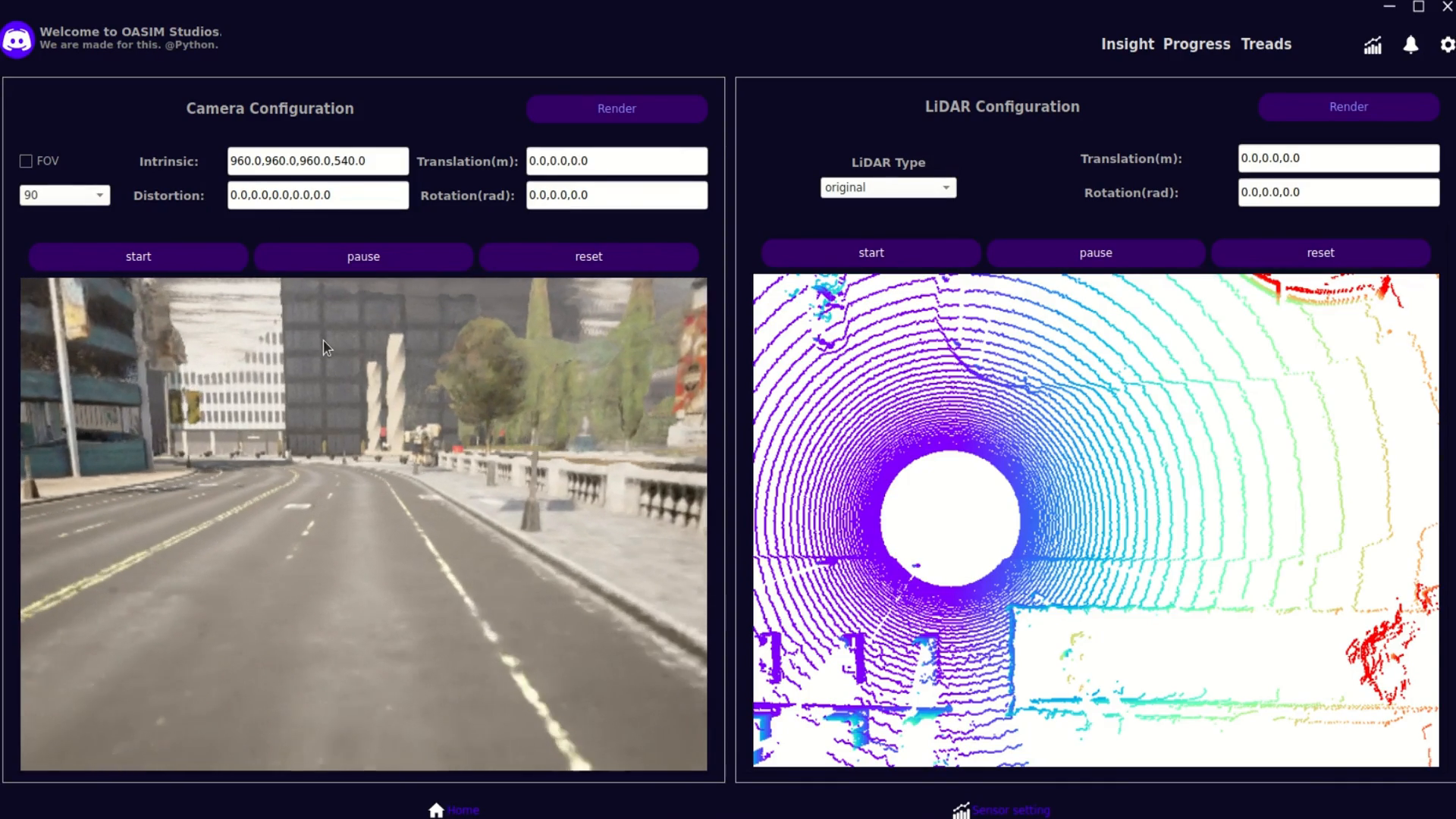}
    \caption{Sensor editing and rendering interface. }
    \label{fig:inteface2}
\end{figure}
\subsection{Downstream Application}
The downstream applications of OASim include several aspects, such as data generation, scenario creation, automatic annotation, and closed-loop training and testing for autonomous driving. As the technology for autonomous driving evolves, sensor solutions and computing platforms become increasingly homogenized, and the technological gap narrows. Consequently, the focus of technological iteration shifts towards the training of algorithmic models, which necessitates substantial data support. OASim is capable of reconstructing and generating vast amounts of data, particularly those uncommon yet crucial edge cases, to train more complex algorithmic models, thereby enhancing the accuracy and reliability of autonomous driving systems. The scenario data required for autonomous driving often constitutes long-tail data, meaning it needs to cover as many driving scenarios as possible. Data closed-loop helps by continuously collecting this long-tail data, aiding autonomous driving systems in better understanding and adapting to a wide variety of complex driving environments. In addition, OASim can provide automatic annotation capabilities, generating data with annotated information. Later stages of OASim can also support tasks such as closed-loop training and testing for autonomous driving.

\section{Experiment}

In this section, we provide experimental result to demonstrate the high-fidelity data generated by the simulator and its ability to edit vehicle trajectories and sensor configurations. 



\textbf{Photorealistic Rendering}


\begin{figure}[htp]
    \centering
    \includegraphics[width=\linewidth]{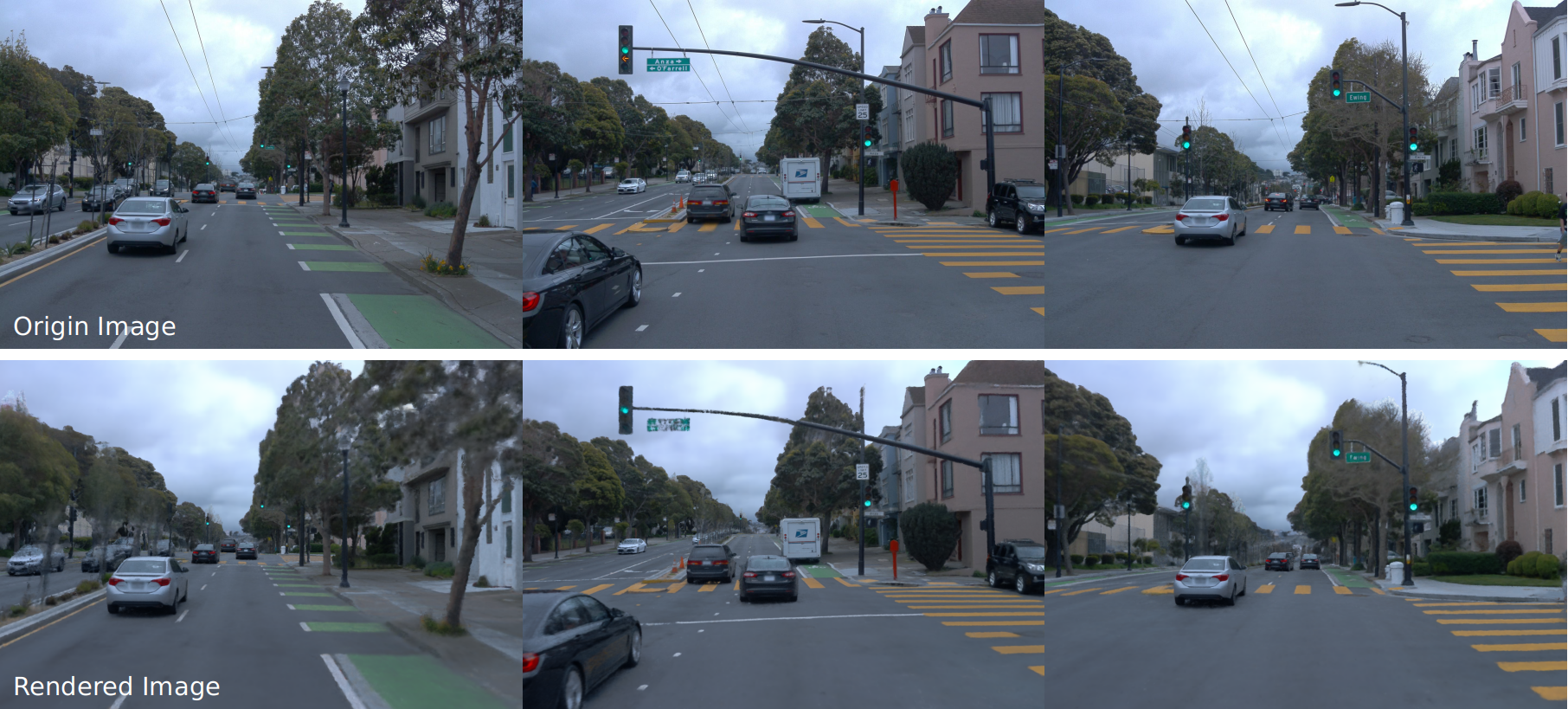}
    \caption{Qualitative image rendering results.}
    \label{fig:exp1}
\end{figure}

\begin{figure}[htp]
    \centering
    \includegraphics[width=\linewidth]{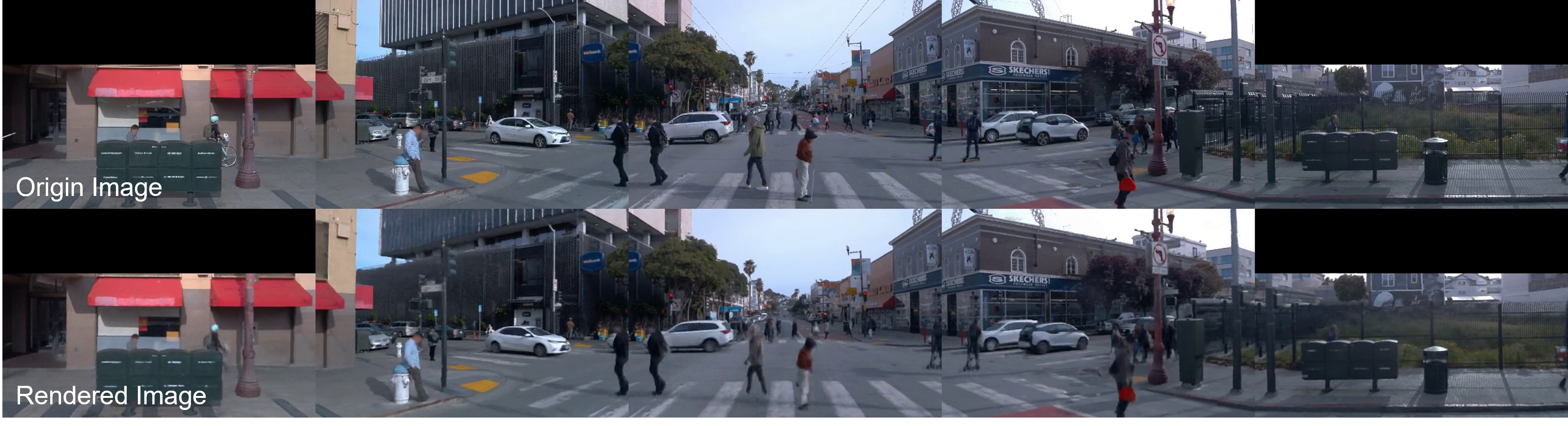}
    \caption{Non-rigid pedestrian rendering results.}
    \label{fig:exp1.1}
\end{figure}

We demonstrate qualitative comparisons of reconstruction with the examples in Waymo Open Dataset. Fig.\ref{fig:exp1} shows three pairs of ground truth image(above) and the rendered image(below). It can be seen that the rendered image is very close to the real picture. Fig.\ref{fig:exp1.1} shows the rendering effect of non-rigid pedestrians

\textbf{Novel View Synthesis} 

\begin{figure}[htp]
    \centering
    \includegraphics[width=\linewidth]{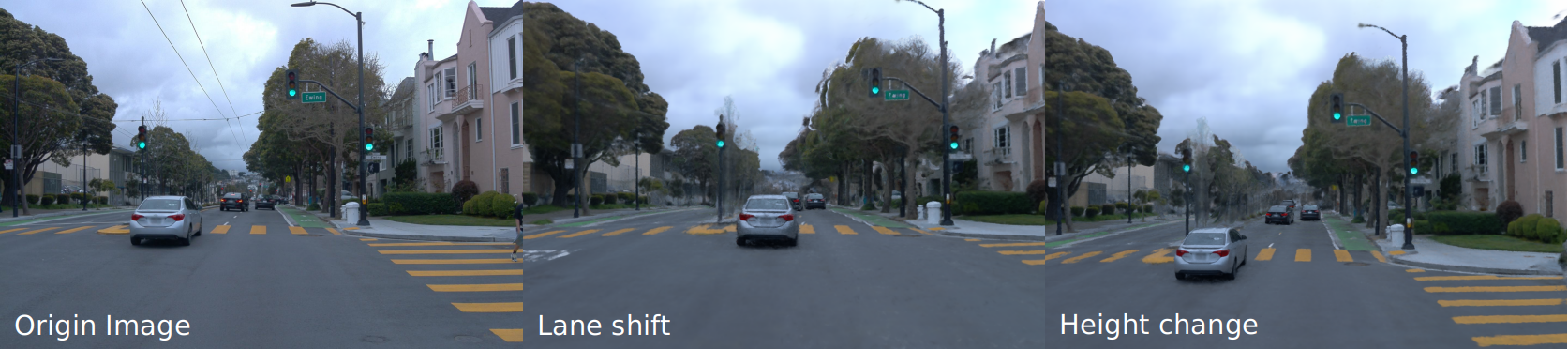}
    \caption{Novel view synthesis results.}
    \label{fig:exp2}
\end{figure}

In order to demonstrate the quality of novel view synthesis, we change the position of the ego vehicle and render the images. The replay image and rendered image is shown in Fig.\ref{fig:exp2}. In the second column, the lane of ego vehicle has changed laterally. In the third column, the camera position has elevated.

\textbf{Diverse Sensor Configuration}

\begin{figure}[htp]
    \centering
    \includegraphics[width=\linewidth]{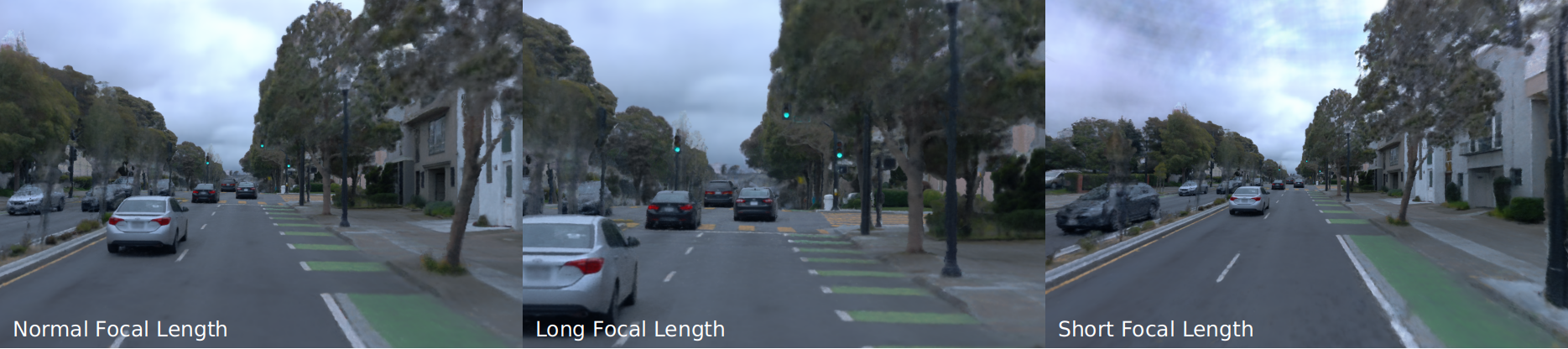}
    \caption{Rendered images of different camera focal length.}
    \label{fig:exp3}
\end{figure}

\begin{figure}[htp]
    \centering
    \includegraphics[width=\linewidth]{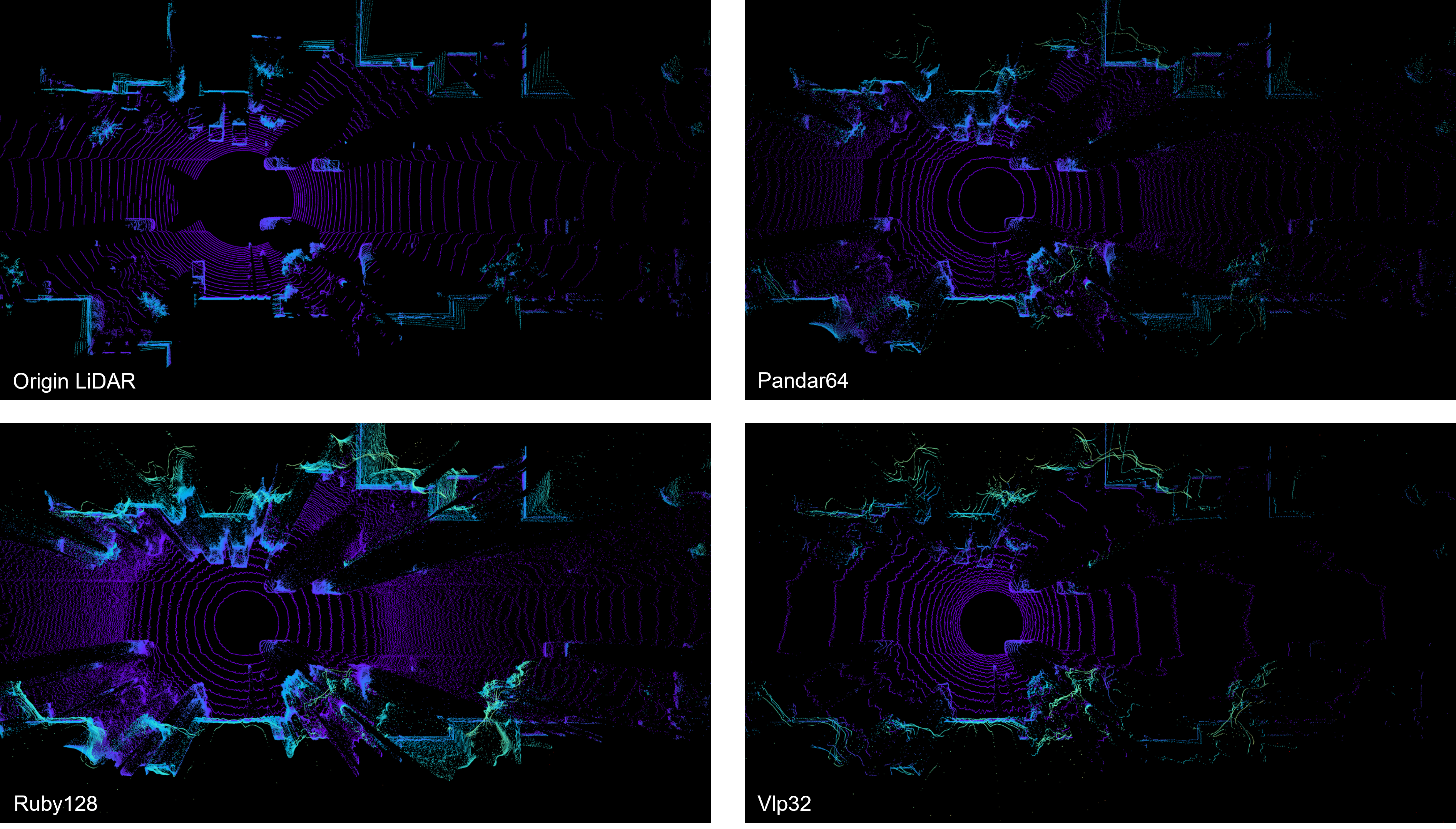}
    \caption{Rendered point clouds of different LiDAR models.}
    \label{fig:exp3.2}
\end{figure}

After reconstructing the environment, we can render the images and LiDAR point clouds with different sensor models. Fig.\ref{fig:exp3} shows the result of cameras with long and short focal lengths in the same scene. Fig.\ref{fig:exp3.2} shows the result of point clouds with different LiDAR models.

\textbf{Diverse Traffic flow simulations}

\begin{figure}[htp]
    \centering
    \includegraphics[width=\linewidth]{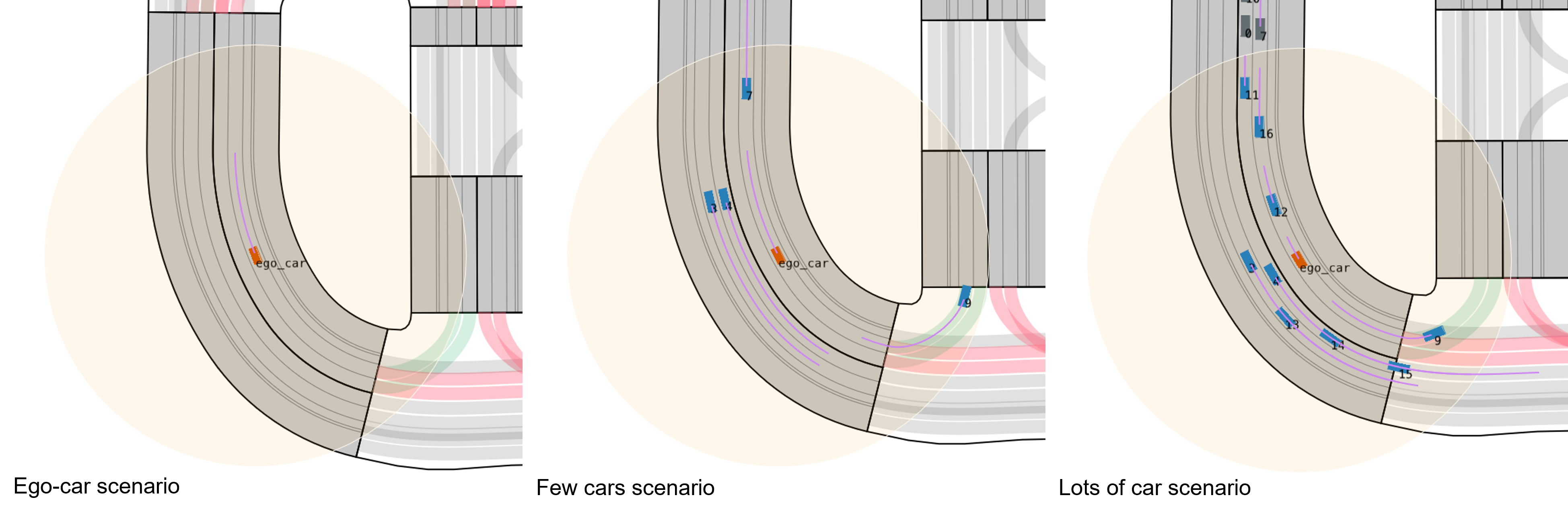}
    \caption{Simulation results of different traffic flow environments.}
    \label{fig:exp4}
\end{figure}

When editing traffic flow, we can simulate different traffic flow scenarios. Fig.\ref{fig:exp4} shows the traffic flow simulation results of ego-car, few cars, and lots of cars.


\section{Conclusion}

We propose OASim, an open-source data generator with realistic rendering capabilities and powerful editing capabilities. From multi-modal data collected in the real world, we can reconstruct the environment and run the vehicle in it freely to generate various data. We can customize corner cases which are of great significance for increasing the capabilities of the model and detecting loopholes in the algorithm. We also provide an interactive interface for users to conveniently edit the settings and visualize the rendering result. The system is readily applicable to outdoor custom dataset and public dataset.


\clearpage


\bibliography{example}  

\end{document}